\def\year{2020}\relax
\documentclass[letterpaper]{article} 
\usepackage{aaai20}  
\usepackage{times}  
\usepackage{helvet} 
\usepackage{courier}  
\usepackage[hyphens]{url}  
\usepackage{graphicx} 
\usepackage{graphicx}  
\usepackage{amssymb}
\usepackage{amsmath}

\usepackage{array}
\usepackage{makecell}
\usepackage{stfloats} 
\usepackage{times}
\usepackage{graphicx}
\usepackage{algorithm}
\usepackage{algorithmic}

\title{Multi-Stage Self-Supervised Learning for Graph Convolutional Networks \\
	on Graphs with Few Labeled Nodes}

\author{Ke Sun\textsuperscript{\rm 1},
	Zhouchen Lin\textsuperscript{\rm 2,5},
	Zhanxing Zhu\textsuperscript{\rm 3,1,4\thanks{Corresponding author}}\\ 
\textsuperscript{1} Center for Data Science, Peking University, China \\
\textsuperscript{2} Key Lab. of Machine Perception (MOE), School of EECS, Peking University, China \\
\textsuperscript{3} School of Mathematical Science, Peking University, China\\ 
\textsuperscript{4} Beijing Institute of Big Data Research (BIBDR)\\ 
\textsuperscript{5} Samsung Research China - Beijing (SRC-B) \\
\{ajksunke, zlin, zhanxing.zhu\}@pku.edu.cn
}

\begin{document}

\maketitle

\begin{abstract}
	Graph Convolutional Networks~(GCNs) play a crucial role in graph learning tasks, however, learning graph embedding with few supervised signals is still a difficult problem. In this paper, we propose a novel training algorithm for Graph Convolutional Network, called Multi-Stage Self-Supervised~(M3S) Training Algorithm, combined with self-supervised learning approach, focusing on improving the generalization performance of GCNs on graphs with few labeled nodes. Firstly, a Multi-Stage Training Framework is provided as the basis of M3S training method. Then we leverage DeepCluster technique, a popular form of self-supervised learning, and design corresponding aligning mechanism on the embedding space to refine the Multi-Stage Training Framework, resulting in M3S Training Algorithm. Finally, extensive experimental results verify the superior performance of our algorithm on graphs with few labeled nodes under different label rates compared with other state-of-the-art approaches.
\end{abstract}

\section{Introduction}

With great expressive power, graphs have been employed as the representation of a wide range of systems across various areas, including social network~\cite{kipf2016semi,hamilton2017inductive}, physical systems~\cite{battaglia2016interaction,sanchez2018graph}, protein-protein interaction networks~\cite{hamaguchi2017knowledge} and knowledge graph~\cite{fout2017protein}.
Recently, research of analyzing graphs with machine learning has been received more and more attention, mainly focusing on node classification~\cite{kipf2016semi}, link prediction~\cite{zhu2016max} and clustering tasks~\cite{fortunato2010community}.

Graph convolution can be regarded as the extension of standard convolution from Euclidean  to non-Euclidean domain. Graph Convolutional Networks~(GCNs)~\cite{kipf2016semi} generalize convolutional neural networks~(CNNs) to graph-structured data from the perspective of spectral theory based on prior works~\cite{bruna2013spectral,defferrard2016convolutional}. GCNs naturally integrate the connectivity patterns and feature attributes of graph-structured data and it has been demonstrated that GCNs and their variants~\cite{hamilton2017inductive,velickovic2017graph,dai2018learning,chen2017stochastic} significantly outperform traditional multi-layer perceptron~(MLP) models and traditional graph embedding approaches~\cite{tang2015line,perozzi2014deepwalk,grover2016node2vec}.

Nevertheless, it is well known that deep neural networks heavily depend on a large amount of labeled data. The requirement of large-scale data might not be met in many real scenarios for graphs with sparse labeled nodes. GCNs and their variants are mainly established on semi-supervised setting where the graph usually has relative plenty of labeled data. However, to the best of our knowledge, there is hardly any work about graphs focusing on weakly supervised setting~\cite{zhou2017brief}, especially learning a classification model with few examples from each class. In addition, the GCNs are usually with shallow architectures due to its intrinsic limitation~\cite{li2018deeper}, thereby restricting the efficient propagation of label signals. To address this issue, \cite{li2018deeper} proposed Co-Training and Self-Training to enlarge training dataset in a boosting-like way. Although these methods can partially improve the performance of GCNs with few labeled data, it is difficult to pick single one consistently efficient algorithm in real applications since these proposed methods~\cite{li2018deeper} perform inconsistently across distinct training sizes.

On the other hand, a recent surge of interest has focused on the self-supervised learning, a popular form of unsupervised learning, which uses pretext tasks to replace the labels annotated by humans by ``pseudo-label'' directly computed from the raw input data. On the basis of the analysis above, there are mainly two issues worthy to explore further. Firstly, since it is hard to change the innate shallow architectures of GCNs, how to design a consistently efficient training algorithm based on GCNs to improve its generalization performance on graphs with few labeled nodes? Secondly, how to leverage the advantage of self-supervised learning approaches based on a large amount of unlabeled data, to refine the performance of proposed training algorithm?

In this paper, we firstly analyze the Symmetric Laplacian Smoothing~\cite{li2018deeper} of GCNs and show that this intrinsic property determines the shallow architectures of GCNs, thus restricting its generalization performance on only few labeled data due to the inefficient propagation of label information. Then we show the layer effect of GCNs on graph with few labeled nodes: to maintain the best generalization, it requires more layers for GCNs with fewer labeled data in order to propagate the weak label signals  more broadly. Further, to overcome the inefficient propagation of label information with few labels for shallow architectures of GCNs, we firstly propose a more general training algorithm of GCNs based on Self-Training~\cite{li2018deeper}, called \textit{Multi-Stage Training  Framework}. Furthermore, we apply DeepCluster~\cite{caron2018deep}, a popular method of self-supervised learning, on the graph embedding process of GCNs and design a novel aligning mechanism on clusters to construct pseudo-labels in classification for each unlabeled data in the embedding space. Next we incorporate DeepCluster approach and the aligning mechanism into the Multi-Stage Training Framework in an elegant way and formally propose \textit{Multi-Stage Self-Supervised~(M3S) Training Algorithm}. Extensive experiments demonstrate that our M3S approach are superior to other state-of-the-art approaches across all the considered graph learning tasks with limited number of labeled nodes. In summary, the contributions of the paper are listed below:
\begin{itemize}
	\item We first probe the existence of Layer Effect of GCNs on graphs with few labeled nodes, revealing that GCNs requires more layers to maintain the performance with lower label rate.
	
	\item We propose an efficient training algorithm, called M3S,  combining the Multi-Stage Training Framework and DeepCluster approach. It exhibits state-of-the-art performance on graphs with low label rates.
	
	\item Our M3S Training Algorithm in fact can provide a more general framework that leverages self-supervised learning approaches to improve multi-stage training framework to design efficient algorithms on learning tasks with only few labeled data.
	
\end{itemize}

\section{Our Approach}
Before introducing our M3S training algorithm, we will firstly elaborate the issue of inefficient propagation of information from limited labeled data due to the essence of symmetric laplacian smoothing of GCNs, which forms the motivation of our work. Then a multi-stage training framework and DeepCluster approach are proposed, respectively, composing the basic components of our M3S algorithm. Finally, we will formally provide multi-stage self-supervised~(M3S) training algorithm in detail, a novel and efficient training method of GCNs focusing on graphs with few labeled nodes.

\subsection{Symmetric Laplacian Smoothing of Graph Convolutional Networks}

In the GCNs model~\cite{kipf2016semi} of semi-supervised classification, the graph embedding $Z$ of nodes with two convolutional layers is formulated as:
\begin{equation} \begin{aligned} 
Z=\text{softmax}(\hat{A} \ \text{ReLU}(\hat{A}XW^{(0)}) W^{(1)}),
\end{aligned} \end{equation}
where $\hat{A}=\tilde{D}^{-\frac{1}{2}}\tilde{A}\tilde{D}^{-\frac{1}{2}}, \tilde{A}=A+I$ and $\tilde{D}$ is the degree matrix of $\tilde{A}$. $X$ and $A$ denote the feature and the adjacent matrix, respectively. $W^{(0)}$ is the input-to-hidden weight matrix and $W^{(1)}$ is the hidden-to-output weight matrix.

Related work~\cite{li2018deeper} pointed out the reason why the GCNs work lies in the Symmetric Laplacian Smoothing of this spectral convolutional type, which is the key for the huge performance gain. We simplify it as follows:

\begin{equation} \begin{aligned} 
{\bf z}_i=\sum_{j}^{}\frac{\tilde{a}_{ij}}{\sqrt{\tilde{d}_i}\sqrt{\tilde{d}_j}}{\bf x}_j \ \ \ (\text{for} \  1\le i \le n),
\end{aligned} \end{equation}
where $n$ is the size of nodes and $\bf z_i$ is the first-layer embedding of node $i$ from input features $\bf x$. Its corresponding matrix formulation is as follows:
\begin{equation} \begin{aligned} 
Z = \tilde{D}^{-\frac{1}{2}}\tilde{A}\tilde{D}^{-\frac{1}{2}}X,
\end{aligned} \end{equation}
where $Z$ is the one-layer embedding matrix of feature matrix $X$. In addition, \cite{li2018deeper} showed that by repeatedly applying Laplacian smoothing many times, the embedding of vertices will finally converge to the proportional to the square root of the vertex degree, thus restricting the enlargement of convolutional layers. 

In this case, a shallow GCN cannot sufficiently propagate the label information to the entire graph with only a few labels, yielding the unsatisfying performance of GCNs on graphs with few labeled nodes. To tackle this deficit of GCNs, we propose an effective training algorithm based on GCNs especially focusing on graphs with only a small number of labels, dispensing with the inconsistent performance of four algorithms proposed in \cite{li2018deeper}.

On the other hand, as shown in Figure~\ref{figure_layers}, the requirement of number of graph convolutional layers for the best performance differs for the different label rates. Concretely speaking, the lower label rate of a graph has, the more graph convolutional layers are required for the purpose of more efficient propagation of label information.

\subsection{Multi-Stage Training Framework}

Inspired by the Self-Training algorithm proposed by \cite{li2018deeper}, working by adding the most confident predictions of each class to the label set, we propose a more general Multi-Stage Training Framework described in Algorithm~\ref{alg_multistage}.

\begin{figure*}[b]
	\centering
	\includegraphics[width=1.0\textwidth]{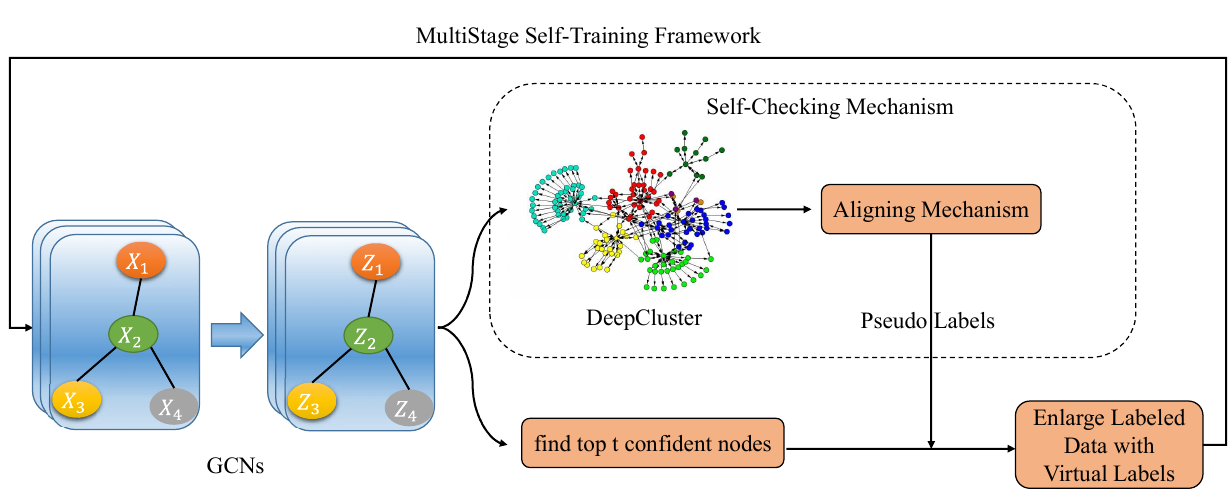}
	\caption{Flow chart of Multi-Stage Self-Supervised~(M3S) Training Algorithm.}
	\label{figure_M3S}
\end{figure*}

In contrast with original Self-Training that explores the most confident nodes and adds them with predicted virtual labels only once, Multi-Stage Training Algorithm executes this process $K$ times. On graphs with limited labels, this algorithm framework repeatedly adds more confident labeled data and facilitates the propagation of label information, resulting in the better performance compared with original approaches.

Nevertheless, the core of Multi-Stage Training Framework lies in the accuracy of selected nodes with virtual labels based on the confidence and thus it is natural to incorporate self-checking mechanism that can guarantee the precision of chosen labeled data.

\begin{algorithm}[htbp]
	\caption{Multi-Stage Training Algorithm}
	\label{alg_multistage}
	\textbf{Input}: Features matrix $X$, adjacent matrix $A$, labeled and unlabeled set $L_0, U_0$, graph convolution network $f_{\theta}$\\
	\textbf{Output}: Graph Embedding $Z=f_{\theta}(X,A)$  
	
	\begin{algorithmic}[1] 
		\STATE Train a fixed number of epoches on the initial labeled and unlabeled set $L_0, U_0$
		\FOR{each stage k}	
		\STATE Sort vertices on confidence in unlabeled set $U_{k-1}$.
		\FOR{each class j}
		\STATE Find the top $t$ vertices in $Z_{i,j}$.
		\STATE Add them to labeled set $L_{k-1}$ with virtual labels $j$.
		\STATE Delete them from unlabeled set $U_{k-1}$.
		\ENDFOR
		\STATE Train a fixed number of epoches on the new labeled and unlabeled set $L_{k}, U_{k}$
		\ENDFOR
		\STATE \textbf{return} Accuracy based on the final $Z$. 
	\end{algorithmic}
\end{algorithm}

\subsection{DeepCluster}

Recently, self-supervised learning~\cite{doersch2015unsupervised}, a popular form of unsupervised learning, shows its power in the field of computer vision, which utilizes pretext tasks to replace the labels annotated by human by ``pseudo-labels''. A neat and effective approach of self-supervised learning is DeepCluster~\cite{caron2018deep} that takes a set of embedding vectors produced by ConvNet $F$ as input and groups them into $k$ distinct clusters based on a geometric criterion.

More concretely, DeepCluster jointly learns a $d \times k$ centroid matrix $C$ and the cluster assignment $y_n$ of each data point $n$ such as image, by solving the following problem:

\begin{equation} \begin{aligned} 
&\min_{C \in \mathcal{R}^{d \times k}} \frac{1}{N} \sum_{n=1}^{N} \min_{y_n \in \{0,1\}^k} \Vert F(x_n)-Cy_n \Vert^2_2 \\ 
& \ \ \ \ \ s.t. \ \ \ \  y_n^T 1_k=1
\end{aligned} \end{equation}

Solving this problem provides a set of optimal assignments $(y_n^*)_{n \le N}$ and a centroid matrix $C^*$. These assignments are then used as pseudo-labels. In particular, DeepCluster alternates between clustering the embedding vectors produced from ConvNet into pseudo-labels and updating parameters of the ConvNet by predicting these pseudo-labels.

For the node classification task in a graph, the representation process can also be viewed as graph embedding~\cite{zhou2018graph}, allowing the DeepCluster as well. Thus, we harness the innate property of graph embedding in GCNs and execute k-means on the embedding vectors to cluster all nodes into distinct categories based on embedding distance. Next, an aligning mechanism is introduced to classify the nodes in each cluster to the nearest class in classification on the embedding space. Finally, the obtained pseudo-labels are leveraged to construct the self-checking mechanism of Multi-Stage Self-Supervised Algorithm as shown in Figure~\ref{figure_M3S}.

\paragraph{Aligning Mechanism} The target of aligning mechanism is to transform the categories in clustering to the classes in classification based on the embedding distance. For each cluster $l$ in unlabeled data after k-means, the computation of aligning mechanism is:
\begin{equation} \begin{aligned} 
c^{(l)} = \mathop{\arg\min}_{m} \Vert v_l-\mu_m \Vert^2,
\end{aligned} \end{equation}
where $\mu_m$ denotes centroids of class $m$ in labeled data, $v_l$ denotes the centroid of cluster $l$ in unlabeled data and $c^{(l)}$ represents the aligned class that has the closest distance from $v_l$ among all centroids of class in the original labeled data. Through the aligning mechanism, we are capable of classifying nodes of each cluster to a specific class in classification and then construct pseudo-labels for all unlabeled nodes according to their embedding distance.

\paragraph{Extension} In fact, DeepCluster is a more general and economical form of constructing self-checking mechanism via embedding distance. The naive self-checking way is to compare the distance of each unlabeled node to centroids of classes in labeled data since distance between each unlabeled data and training centroids is a more precise measure than the class centriods of unlabeled data. However, when the number of clusters is equivalent to the amount of all unlabeled nodes, our self-checking mechanism via DeepCluster is the same as the naive way. Considering the expensive computation of the naive self-checking, DeepCluster performs more efficiently and flexibly in the selection of number of clusters.

\begin{algorithm}[t!]
	\caption{M3S Training Algorithm}
	\label{alg:M3S-GCN}
	\textbf{Input}: Features Matrix $X$, adjacent matrix $A$, labeled and unlabeled set $L_0, U_0$, graph convolution network $f_{\theta}$.\\
	\textbf{Output}: Graph Embedding $Z=f_{\theta}(X,A)$  
	
	\begin{algorithmic}[1] 
		\STATE Train a fixed number of epoches on the initial labeled and unlabeled sets $L_0, U_0$.
		\FOR{each stage k}
		\STATE \textbf{\% Step 1: Deep Clustering}	
		\STATE Execute K-means based on embedding $Z$ of all data and obtain pseudo labels of each data point for clustering.
		\STATE \textbf{\% Step 2: Aligning Mechanism}	
		\STATE Compute centroids $\mu_m$ of each class $m$ in labeled data. \\
		\STATE Compute centroids $v_l$ of each cluster $l$ in unlabeled data.\\
		\FOR{each cluster $l$ of unlabeled set}
		\STATE Align label of $l_{th}$ cluster on the embedding space.
		$$c^{(l)} = \mathop{\arg\min}_{m} \Vert v_l-\mu_m \Vert^2 $$\\
		\STATE Set unlabeled data in $l_{th}$ cluster with pseudo label $c^{(l)}$.
		\ENDFOR
		\STATE \textbf{\% Step 3: Self-Training}	
		\STATE Sort vertices according to the confidence in unlabeled set $U_{k-1}$.
		\FOR{each class j}
		\STATE Find the top $t$ vertices in $Z_{i,j}$.
		\FOR{each vertice of selected $t$ vertices}
		\IF{pseudo label $c$ of the vertice equals j}
		\STATE Add it to labeled set $L_{k-1}$ with virtual label $j$.
		\STATE Delete it from unlabeled set $U_{k-1}$.
		\ENDIF
		\ENDFOR
		\ENDFOR
		\STATE Train a fixed number of epoches on the new labeled and unlabeled set $L_{k}, U_{k}$
		\ENDFOR
		\STATE \textbf{return} Accuracy based on the final $Z$. 
	\end{algorithmic}
\end{algorithm}

\subsection{M3S Training Algorithm}

In this section, we will formally present our Multi-Stage Self-Supservised~(M3S) Training Algorithm, a novel training method on GCN aiming at addressing the inefficient propagation of label information on graphs with few labeled nodes. The flow chart of our approach is illustrated in Figure~\ref{figure_M3S}.

The crucial part of M3S Training Algorithm compared with Multi-Stage Training is additionally utilizing the information of embedding distance to check the accuracy of selected nodes with virtual labels from Self-Training based on the confidence. Specifically speaking, M3S Training Algorithm elegantly combines DeepCluster self-checking mechanism with Multi-Stage Training Framework to choose nodes with more precise virtual labels in an efficient way. We provide a detailed description of M3S approach in Algorithm~\ref{alg:M3S-GCN}.

For M3S Training Algorithm, firstly we train a GCN model on an initial dataset to obtain meaningful embedding vectors. Then we perform DeepCluster on the embedding vectors of all nodes to acquire their clustering labels. Furthermore, we align their labels of each cluster based on the embedding distance to attain the pseudo-label of each unlabeled node.  In the following Self-Training process, for the selected top confident nodes of each class, we perform self-checking based on pseudo-labels to guarantee they belong to the same class in the embedding space, then add the filtered nodes to the labeled set and execute a new stage Self-Training.

\paragraph{Avoiding Trivial Solutions} It should be noted that the categorically balanced labeled set plays an important role on graphs with low label rate. In addition, DeepCluster tends to be caught in trivial solutions that actually exist in various methods that jointly learns a discriminative classifier and the labels~\cite{caron2018deep}. Highly unbalanced data of per class is a typical trivial solution of DeepCluster, which hinders the generalization performance with few supervised signals. In this paper we provide a simple and elegant solution by enlarging the number of clusters in K-means. For the one hand, setting more clusters allows higher probability of being evenly classified to all categories. For the other hand, it contributes to more precise computation in embedding distance from the perspective of extension of DeepCluster self-checking mechanism. These are dicussed in the experimental part.

\section{Experiments}

\begin{figure*}[t!]
	\begin{center}
		\centering
		\includegraphics[width=0.8\textwidth]{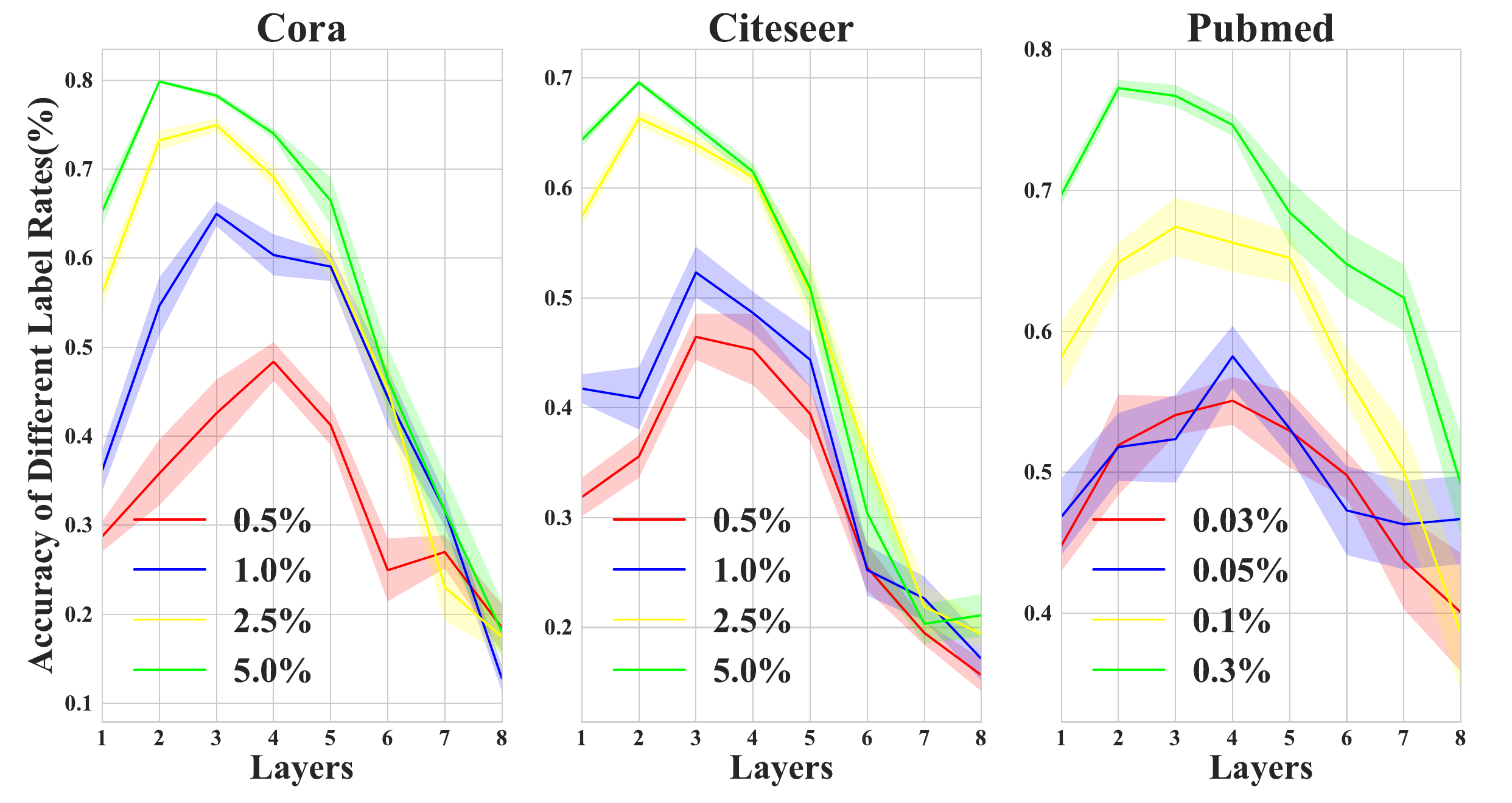}
		\caption{The change of accuracy for GCNs model with different layers under different label rates.}
		\label{figure_layers}
	\end{center}
\end{figure*}

In this section we conduct extensive experiments to demonstrate the effectiveness of our proposed M3S Algorithm on graphs with few labeled nodes. For the graph dataset, we select the three commonly used citation networks: CiteSeer, Cora and PubMed~\cite{sen2008collective}. Dateset statistics are summarized in Table~\ref{table_dataset}.

As for the baselines, we opt the Label Propagation~(LP) \cite{wu2012learning} using
ParWalks; Graph Convolutional Networks~(GCNs)~\cite{kipf2016semi}; Self-Training, Co-Training, Union and Intersection~\cite{li2018deeper} all based on the confidence of prediction. On graphs with low label rates, we compare both our Multi-Stage Training Framework and M3S Algorithm with other state-of-the-art approaches by changing the label rate for each dataset. We report the mean accuracy of 10 runs in all result tables to make fair comparison. Our implementationm, including the splitting of train and test datasets, adapts from original version in~\cite{li2018deeper}.

\begin{table}[htbp]
	\centering
	\begin{tabular}{lrrrrr}
		\hline
		\textbf{Dateset}  & \textbf{Nodes} & \textbf{Edges} & \textbf{Classes} & \textbf{Features} & \makecell{\bf Label \\ \bf Rate} \\
		\hline
		CiteSeer &    3327&    4732&    6&    3703&  3.6\%\\
		Cora     &    2708&    5429&    7&    1433&  5.2\%\\
		PubMed   &   19717&   44338&    3&     500&  0.3\%\\
		\hline
	\end{tabular}
	\caption{Dateset statistics}
	\label{table_dataset}
\end{table}

\subsection{Layer Effect on Graphs with Few Labeled Nodes}\label{layereffect}

\begin{figure*}[t!]
	\begin{center}
		\centering
		\includegraphics[width=.72\textwidth]{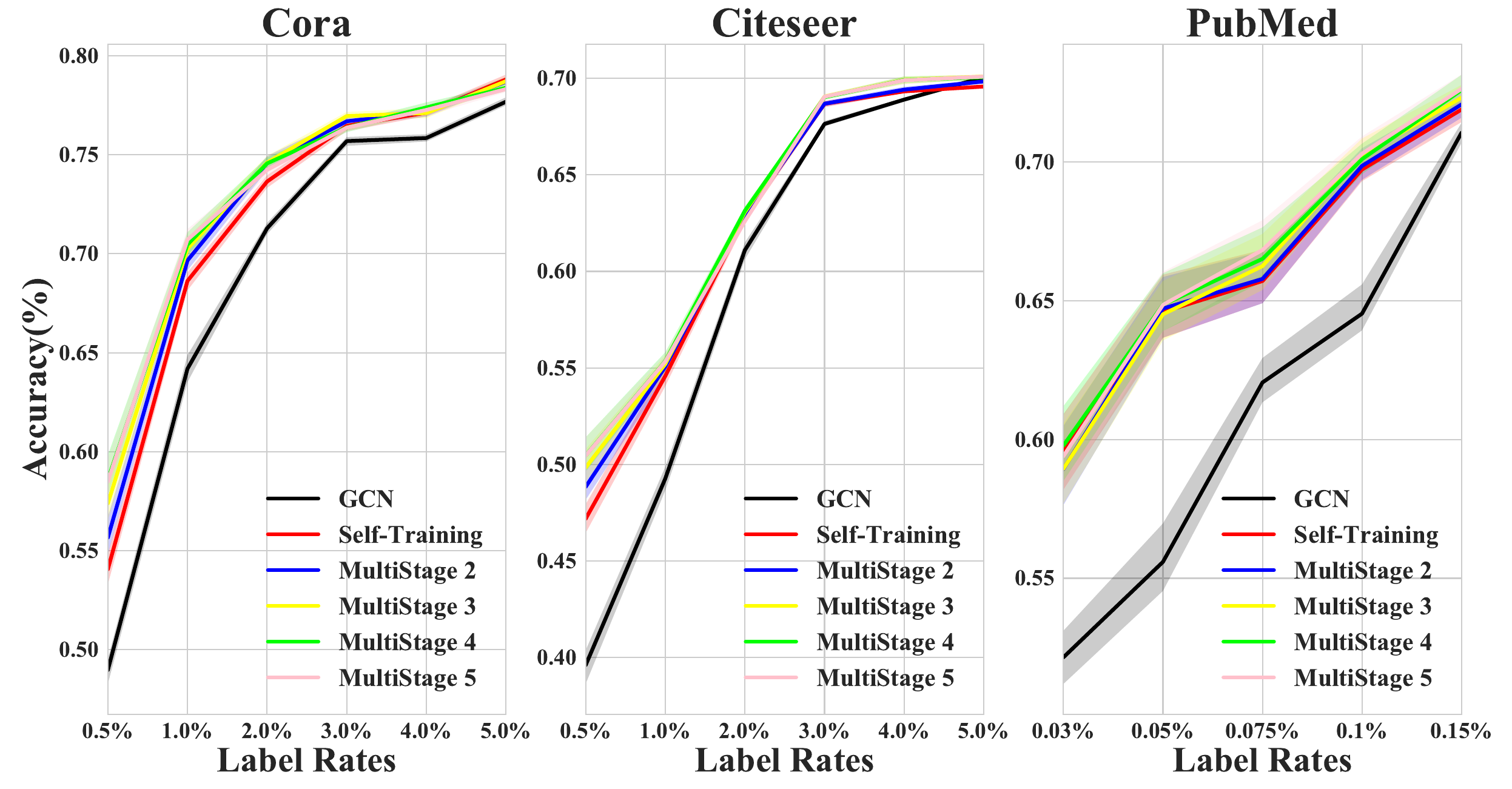}
		\caption{Multi-Stage Training vs Self-training.}
		\label{figure_multistage}
	\end{center}
\end{figure*}

Before comparing our algorithms with other methods, we point out the layer effect of GCNs for different label rates: to maintain the best performance, a GCN model in semi-supervised task with a lower label rate requires more graph convolutional layers.

Figure~\ref{figure_layers} presents some empirical evidence to demonstrate the layer effect on graphs with few labels. We test the performance of GCNs with different layers in distinct label rates in Figure~\ref{figure_layers} and it is apparent to note that the number of layer under the best performance exhibits a descending trend as the label rate increases.

The existence of layer effect demonstrates the urge of propagation of label information by stacking more convolutional layers. In the original GCNs~\cite{kipf2016semi}, the authors argued to apply two graph convolutional layers for standard node classification tasks. However, due to the existence of Layer Effect, we are expected to choose proper number of layers especially on graphs with low label rates. In the following experiments, we all choose the best number of layer to compare the best performance for each method.

\subsection{Performance of Multi-Stage Training Framework}

To gain a better understanding of the advantage of Multi-Stage Training Framework, we make an extensive comparison between Multi-Stage Framework of different stages with the Self-Training approach under different label rates.

From Figure~\ref{figure_multistage}, it is easy to observe that all self-training methods outperform the original GCNs with a large margin, especially when the graph has low label rate, which usually happens in real applications. In addition, Multi-Stage Training is superior to traditional Self-Training especially when there are fewer labeled nodes and more stages are inclined to bring more improvement. Nevertheless, the discrepancy between the Multi-Stage Training algorithm and Self-Training algorithm narrows down as the label rate increases. Moreover, the improvement of all self-training methods over GCNs diminishes as well with the increasing of label rate. As for the reason, we argue that with the enlargement of labeled nodes, the accuracy of the learned GCN model also increases, while the accuracy of explored nodes via self-training tends to approach the accuracy of current GCN, resulting in the diminishment of improvement. However, the limited precision of selected nodes only based on the confidence of prediction is just what M3S Training Algorithm is devoted to improve.

\subsection{Performance of M3S Training Algorithm}

In this section, we conduct experiments by comparing Multi-Stage Self-Training Algorithm and M3S Training Algorithm with other state-of-the-art approaches under different label rates across the three datasets. 

\paragraph{Experimental Setup} All the results are the mean accuracy of 10 runs and the number of clusters in DeepCluster is fixed 200 for all datasets to avoid trivial solutions. We select the best number of layers for different label rates. In particular, the best layer in Cora and CiteSeer is 4,3,3,2,2 and 3,3,3,2,2 respectively for 0.5\%,1\%,2\%,3\%,4\% label rates and fixed 4 for 0.03\%,0.05\%,0.1\% label rates on PubMed. The number of epochs of each stage in Multi-Stage Training Framework, M3S and other approaches is set as 200. For all methods involved in GCNs, we use the same hyper-parameters as in \cite{kipf2016semi}: learning rate of 0.01, 0.5 dropout rate, $5 \times 10^{-4} \ L_2$ regularization weight, and 16 hidden units without validation set for fair comparison~\cite{li2018deeper}. For the option of $K$ stages, we view it as a hyper-parameter. For CiteSeer dataset we fix $K=3$ and for PubMed dataset we fix $K=4$, in which the result of our proposed algorithms have already outperformed other approaches easily. For Cora dataset we choose $K$ as 5,4,4,2,2 as the training size increases, since higher label rate usually matches with a smaller $K$.

\begin{table}[b!]
	\centering{\textbf{Cora Dataset}}
	\begin{tabular}{lrrrrr}
		\hline
		\textbf{Label Rate}  & 0.5\% & 1\% & 2\% & 3\% & 4\% \\
		\hline
		\textbf{LP}            &    57.6&    61.0&    63.5&    64.3&    65.7\\
		\textbf{GCN}           &    50.6&    58.4&    70.0&    75.7&    76.5\\
		\textbf{Co-training}   &    53.9&    57.0&    69.7&    74.8&    75.6\\
		\textbf{Self-training} &    56.8&    60.4&    71.7&    76.8&    77.7\\
		\textbf{Union}         &    55.3&    60.0&    71.7&    77.0&    77.5\\
		\textbf{Intersection}  &    50.6&    60.4&    70.0&    74.6&    76.0\\
		\hline
		\textbf{MultiStage}    &    61.1&    63.7&    74.4&    76.1&    77.2\\
		\textbf{M3S}       &\bf 61.5&\bf 67.2&\bf 75.6&\bf 77.8&\bf 78.0\\
		\hline
	\end{tabular}
	\caption{Classification Accuracy on Cora.}
	\label{table_weakly_cora}
\end{table}

\begin{table}[b!]
	\centering{\textbf{CiteSeer Dataset}}
	\begin{tabular}{lrrrrr}
		\hline
		\textbf{Label Rate}  & 0.5\% & 1\% & 2\% & 3\% & 4\%\\
		\hline
		\textbf{LP}           &    37.7&    41.6&    41.9&    44.4&    44.8\\
		\textbf{GCN}          &    44.8&    54.7&    61.2&    67.0&    69.0\\
		\textbf{Co-training}  &    42.0&    50.0&    58.3&    64.7&    65.3\\
		\textbf{Self-training}&    51.4&    57.1&    64.1&    67.8&    68.8\\
		\textbf{Union}        &    48.5&    52.6&    61.8&    66.4&    66.7\\
		\textbf{Intersection} &    51.3&    61.1&    63.0&    69.5&    70.0\\
		\hline
		\textbf{MultiStage}   &    53.0&    57.8&    63.8&    68.0&    69.0\\
		\textbf{M3S}      &\bf 56.1&\bf 62.1&\bf 66.4&\bf 70.3&\bf 70.5\\
		\hline
	\end{tabular}
	\caption{Classification Accuracy on CiteSeer.}
	\label{table_weakly_citeseer}
\end{table}

\begin{table}[b!]
	\centering\textbf{PubMed Dataset}
	\begin{tabular}{lrrr}
		\hline
		\textbf{Label Rate}  & 0.03\% & 0.05\%& 0.1\%\\
		\hline
		\textbf{LP}           &    58.3&    61.3&    63.8\\
		\textbf{GCN}          &    51.1&    58.0&    67.5\\
		\textbf{Co-training}  &    55.5&    61.6&    67.8\\
		\textbf{Self-training}&    56.3&    63.6&    70.0\\
		\textbf{Union}        &    57.2&    64.3&    70.0\\
		\textbf{Intersection} &    55.0&    58.2&    67.0\\
		\hline
		\textbf{MultiStage}   &    57.4&    64.3&    70.2\\
		\textbf{M3S}          &\bf 59.2&\bf 64.4&\bf 70.6\\
		\hline
	\end{tabular}
	\caption{Classification Accuracy on PubMed.}
	\label{table_weakly_pubmed}
\end{table}

\begin{figure*}[t!]
	\begin{center}
		\centering
		\includegraphics[width=.75\textwidth]{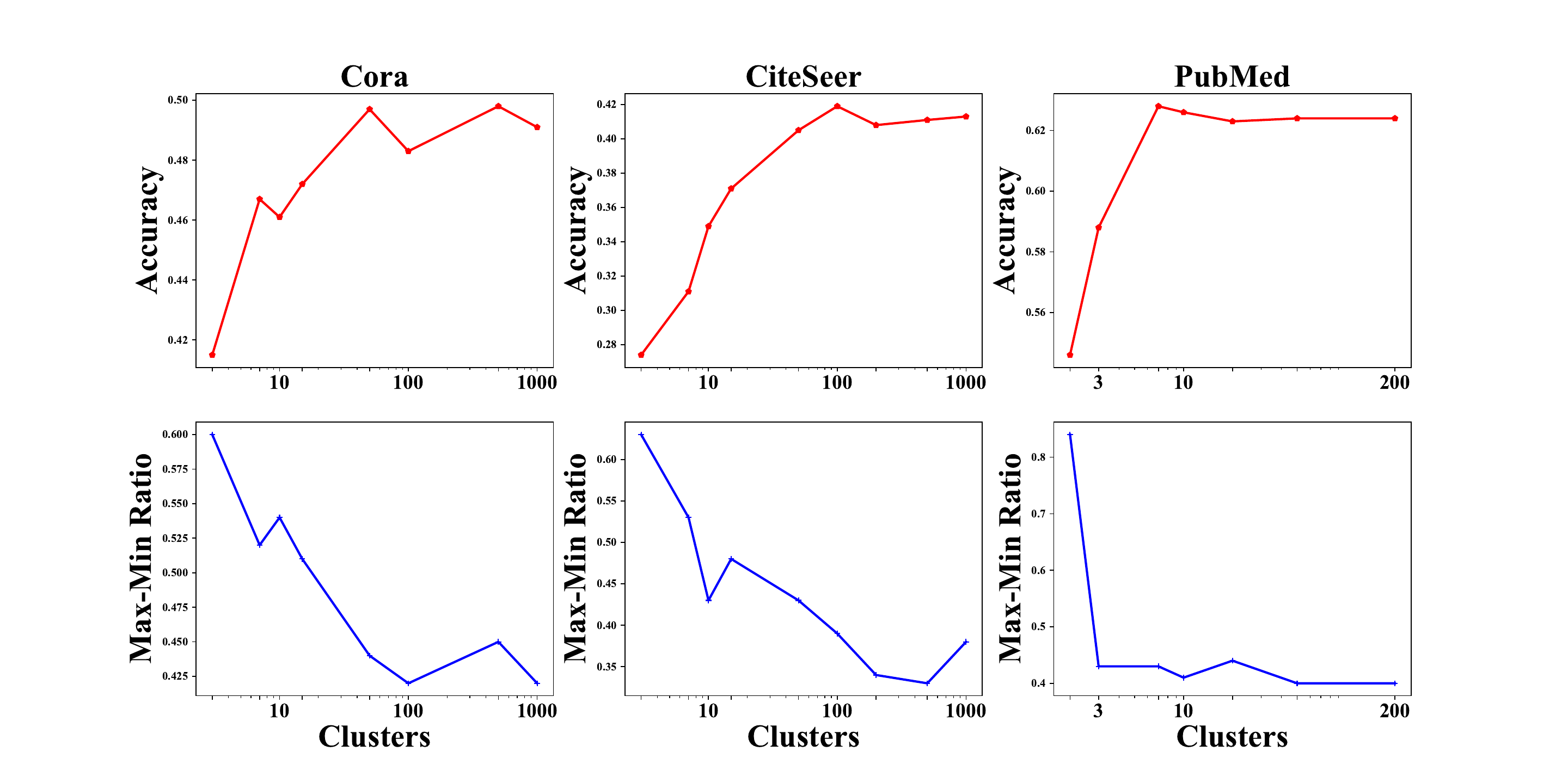}
		\caption{Relation between Accuracy and Max-Min Ratio with the increasing of Clusters $K$. All values are the mean accuracy or max-min ratio of 10 runs.}
		\label{figure_clusters}
	\end{center}
\end{figure*}

Results shown in Tables~\ref{table_weakly_cora},~\ref{table_weakly_citeseer} and~\ref{table_weakly_pubmed} verify the effectiveness of our M3S Training Algorithm, consistently outperforming other state-of-the-art approaches to a large margin on a wide range of label rates across the three datasets. More specifically, we make four observations from the results:

\begin{itemize}
	\item It is apparent to note that the performance of GCN significantly declines when the labeled data is scarce due to the inefficient propagation of label information. For instance, on Cora and PubMed datasets, the performance of GCN is even inferior to Label Propagation~(LP) when the training size is relative small.
	\item Previous state-of-the-art algorithms, namely Co-training, Self-training, Union and Intersection exhibit inconsistent performance compared with GCNs, thus it is hard to employ one single algorithm from them in real scenarios.
	\item Multi-Stage Training Framework tends to be superior to Self-Training especially on fewer labeled data, demonstrating the effectiveness of this framework on graphs with few labeled nodes.
	\item M3S Training Algorithm leverages both the advantage of Multi-Stage Training Framework and self-checking mechanism constructed by DeepCluster, consistently outperforming other state-of-the-art approaches on all label rates. Additionally, it turns out that the lower label rate the graph has, the larger improvement of M3S Training Algorithm can produce, perfectly adapting on graphs with few labeled nodes.
\end{itemize}

\paragraph{Sensitivity Analysis of Number of Clusters} Sensitivity analysis of number of clusters is regarded as the extensive discussion of our M3S Training Algorithm, where we present the influence of number of clusters in DeepCluster on the balance of each class and the final performance of GCN. We leverage ``Max-Min Ratio'' to measure the balance level of each class, which is computed by the subtraction between max ratio and min ratio of categories of unlabeled data after the aligning mechanism, and the lower ``Max-Min Ratio'' represents the higher balance level of categories. We choose two labeled nodes of each class across three datasets. As shown in Figure~\ref{figure_clusters} where each column presents the change of a specific dataset, with the increasing of number of clusters, categories tend to be more balanced until the number of clusters is large enough, facilitating the final performance of M3S Training Algorithm. These results empirically demonstrate that more clusters are beneficial to avoid trivial solutions in DeepCluster, thus enhancing the performance of our method.

\section{Discussions}
Although in this work we employ only one kind of self-supervised approach on the graph learning task, the introduction of self-checking mechanism constructed by DeepCluster in fact provides a more general framework on weakly supervised signals for a wide range of data types. On the one hand, it is worthy of exploring the avenue to utilize the pseudo-labels produced by self-supervised learning more efficiently on few supervised labels, for instance, designing new aligning mechanism or applying better self-supervised learning approach. On the other hand, how to extend similar algorithm combined with self-supervised learning methods to other machine learning task such as image classification and sentence classification, requires more endeavours in the future.

\section{Conclusion}
In this paper, we firstly clarify the Layer Effect of GCNs on graphs with few labeled nodes, demonstrating that it is expected to stack more layers to facilitate the propagation of label information with lower label rate. Then we propose Multi-Stage Training Algorithm Framework on the basis of Self-Training, adding confident data with virtual labels to the labeled set to enlarge the training set. In addition, we apply DeepCluster on the graph embedding process of GCNs and design a novel aligning mechanism to construct self-checking mechanism to improve MultiStage Training Framework. Our final proposed approach, M3S Training Algorithm, outperforms other state-of-the-art methods with different label rates across all the considered graphs with few labeled nodes. Overall, M3S Training Algorithm is a novel and efficient algorithm focusing on graphs with few labeled nodes.

\section{Acknowledgment}
Z. Lin is supported by NSF China (grant no.s 61625301 and 61731018), Major Scientific Research Project of Zhejiang Lab (grant no.s 2019KB0AC01 and 2019KB0AB02), and Beijing Academy of Artificial Intelligence. Zhanxing Zhu is supported in part by National Natural Science Foundation of China (No.61806009), Beijing Natural Science Foundation~(No. 4184090) and Beijing Academy of Artificial Intelligence~(BAAI).

\bibliographystyle{aaai} 
\bibliography{1364M3S}

\end{document}